\documentclass[runningheads]{llncs}

\usepackage{graphicx}
\usepackage{amsmath}
\usepackage{amssymb}

\usepackage{array}
\usepackage{booktabs}
\usepackage{multirow}
\usepackage{makecell}
\usepackage{ulem}
\usepackage[misc]{ifsym}

\usepackage{url}
\usepackage{multirow}
\usepackage{booktabs}
\usepackage{pifont}
\usepackage[dvipsnames]{xcolor}
\usepackage{graphicx}
\usepackage{amsmath}
\usepackage{algorithm}
\usepackage{mathtools}
\usepackage{algorithm}
\usepackage{algpseudocode}
\usepackage{listings}
\usepackage{wrapfig}
\usepackage{colortbl}
\usepackage{array}

\usepackage{rotating}
\usepackage{makecell}
\usepackage{color}

\definecolor{top1}{RGB}{255,179,179}
\definecolor{top2}{RGB}{255,217,179}
\definecolor{top3}{RGB}{255,255,179}
\definecolor{dino}{RGB}{249,231,227}

\usepackage{xcolor,pifont}
\newcommand*\colourcheck[1]{%
  \expandafter\newcommand\csname #1check\endcsname{\textcolor{#1}{\ding{52}}}%
}
\colourcheck{blue}
\colourcheck{green}
\colourcheck{red}
\usepackage{makecell}

\newcommand{\model}{\textit{Endora}}

\newcommand{\da}{Colonoscopic}
\newcommand{\db}{Kvasir-Capsule}
\newcommand{\dc}{CholecTriplet}

\newcommand{\dcS}{Cholec}

\newcommand{\daCC}{\cite{mesejo2016computer}}
\newcommand{\dbCC}{\cite{borgli2020hyperkvasir}}
\newcommand{\dcCC}{\cite{nwoye2022rendezvous}}

\newcommand{\daC}{Colonoscopic~\cite{mesejo2016computer}}
\newcommand{\dbC}{Kvasir-Capsule~\cite{borgli2020hyperkvasir}}
\newcommand{\dcC}{CholecTriplet~\cite{nwoye2022rendezvous}}

\newcommand{\maC}{StyleGAN-V~\cite{skorokhodov2022stylegan}}
\newcommand{\mbC}{MoStGAN-V~\cite{shen2023mostgan}}
\newcommand{\mcC}{LVDM~\cite{he2023latent}}

\newcommand{\dino}{DINO}
\newcommand{\dinoC}{DINO~\cite{caron2021emerging}}

\newcommand{\Mat}{\boldsymbol}

\DeclareMathOperator{\mean}{\mathbb{E}}

\newcommand{\cmark}{\ding{51}}
\newcommand{\xmark}{\ding{55}}
\newcommand{\gcheck}{{\color{ForestGreen}\cmark}}
\newcommand{\redcross}{{\color{BrickRed}\xmark}}

\usepackage[T1]{fontenc}
\usepackage{adjustbox}
\usepackage{multirow}
\usepackage{mathtools}
\usepackage{graphicx}
\usepackage[dvipsnames]{xcolor}
\usepackage[title]{appendix}%

\usepackage[pagebackref=true,breaklinks=true,colorlinks,bookmarks=false]{hyperref}
\begin{document}

\title{
\begin{minipage}{0.1\textwidth}
\includegraphics[width=\linewidth]{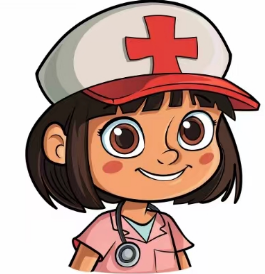}
\end{minipage}
\model: Video Generation Models as \\ Endoscopy Simulators
}
\titlerunning{\model: Video Generation Models as Endoscopy Simulators}
\author{
\textbf{
Chenxin Li\textsuperscript{1}\thanks{Equal contribution}, 
Hengyu Liu\textsuperscript{1$\star$}, 
Yifan Liu\textsuperscript{1$\star$}, 
Brandon Y. Feng\textsuperscript{2}, 
Wuyang Li\textsuperscript{1}, 
Xinyu Liu\textsuperscript{1}, 
Zhen Chen\textsuperscript{3},
Jing Shao\textsuperscript{4},
Yixuan Yuan\textsuperscript{1}$^{(\textrm{\Letter})}$
}
}

\institute{
{\textsuperscript{1} The Chinese University of Hong Kong},  {\textsuperscript{2} Massachusetts Institute of Technology},\\ {\textsuperscript{3} Centre for Artificial Intelligence and Robotics, Hong Kong },
{\textsuperscript{4}Shanghai AI Lab}
}

\authorrunning{C. Li et al.}

\maketitle

\begin{abstract}
Generative models hold promise for revolutionizing medical education, robot-assisted surgery, and data augmentation for machine learning. 
Despite progress in generating 2D medical images, the complex domain of clinical video generation has largely remained untapped.
This paper introduces \model, an innovative approach to generate medical videos that simulate clinical endoscopy scenes.
We present a novel generative model design that integrates a meticulously crafted spatial-temporal video transformer with advanced 2D vision foundation model priors, explicitly modeling spatial-temporal dynamics during video generation. 
We also pioneer the first public benchmark for endoscopy simulation with video generation models, adapting existing state-of-the-art methods for this endeavor.
\model~demonstrates exceptional visual quality in generating endoscopy videos, surpassing state-of-the-art methods in extensive testing.
Moreover, we explore how this endoscopy simulator can empower downstream video analysis tasks and even generate 3D medical scenes with multi-view consistency. 
In a nutshell, \model~marks a notable breakthrough in the deployment of generative AI for clinical endoscopy research, setting a substantial stage for further advances in medical content generation.
For more details, please visit our project page: \url{https://endora-medvidgen.github.io/}.

\end{abstract}
\section{Introduction}

Recent strides in generative AI have sparked significant interest across medical disciplines~\cite{thirunavukarasu2023large,azizi2021can,hu2023synthetic,li2022domain}, pushing the frontiers of computer-aided diagnostics to new heights~\cite{masood2013computer,masood2013computer}.
Amidst a broad array of endeavors in medical AI---ranging from visual question answering~\cite{ben2019vqa,zhan2020medical} and text summarizing~\cite{mishra2014text,kieuvongngam2020automatic} to image reconstruction~\cite{zhang2020review,zeng2010medical,zhang2021generator} and translation~\cite{armanious2020medgan,ozbey2023unsupervised,chen2023generative,liu2024endogaussian}, and even mixed reality for surgical assistance~\cite{lungu2021review,gregory2018surgery}---we venture into uncharted territories and ask: {
Can we generate dynamic, realistic, and complex content like clinical endoscopy videos?
}

Endoscopy is a field at the forefront for advances in gastrointestinal disease diagnosis, minimally invasive procedures, and robotic surgeries~\cite{misawa2021development}. 
Despite its critical role, endoscopic research and training are hindered by the scarcity of visual data, as capturing images inside the body with small endoscopes is inherently difficult. The dire need for a diverse and quality-rich collection of clinical endoscopy videos~\cite{wang2023foundation} underscores the urgency for breakthroughs in medical generative AI. 
We aim to build a powerful endoscopy video simulator and create an extensive array of high-quality endoscopy videos, thus enriching the resources available for medical professionals~\cite{cooke2010educating} and improving training data for surgical robots and AI algorithms.
This exciting venture prompts us to probe deeper into several research questions:
\ding{202} \textbf{Establishing Video Benchmarks}: Medical imaging and text have established benchmark applications, such as automated text report generation and image reconstruction~\cite{mishra2014text,zhang2020review}. Can we extend this success to medical videos and properly benchmark endoscopy simulation quality?
\ding{203} \textbf{Spatial-Temporal Modeling}: While current methods are effective in generating realistic 2D clinical images by generative adversarial networks (GANs)~\cite{yi2019generative} and diffusion models~\cite{kazerouni2022diffusion}, the dynamic nature of endoscopy videos, rich with spatial-temporal correlation, poses a significant challenge. Can our models effectively simulate the intricacies of real-life surgical procedures?

\begin{figure}[t]
 \centering
  \includegraphics[width=1.0\linewidth]{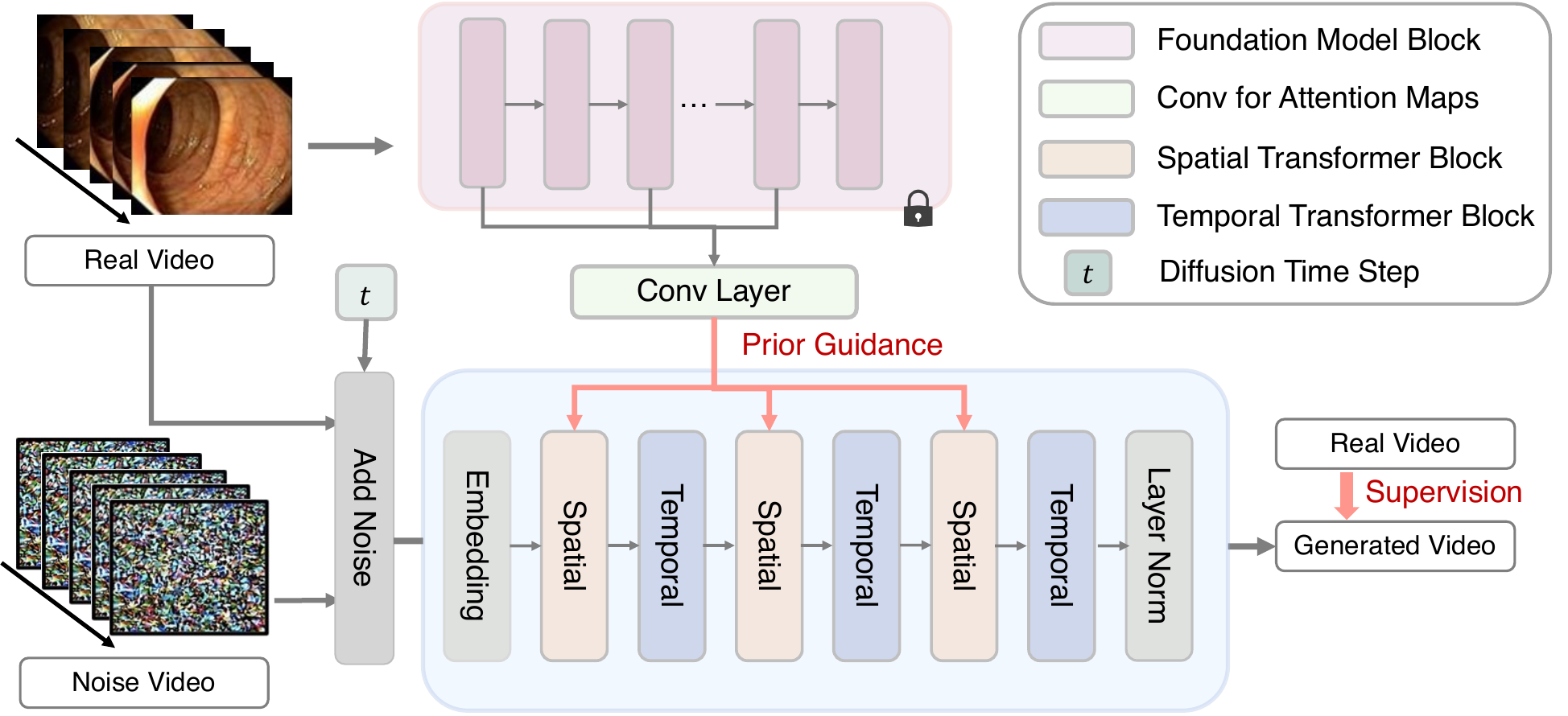}
  \vspace{0.2em}
 \caption{
  \textbf{\model~Training Overview.} 
  Starting from the noised input video sequences, the diffusion model iteratively removes the noise and recover the clean sequence. The long-range spatial-temporal dynamics is modeled by an interlaced cascading of several spatial-temporal Transformer blocks.
  We further instill the prior from 2D vision foundation model (\dinoC) to guide feature extraction.
 } 
 \label{fig:ppline}
\end{figure}

Driven by these questions, 
we formulate a framework to generate spatially and temporally coherent and plausible endoscopy videos to synthesize realistic clinical scenes. 
Marking a departure from traditional approaches to medical content generation that primarily deal with textual and 2D image data, we aspire to set \textbf{a holistic benchmark} for future explorations in video generation models within the medical domain.
In particular, by meticulously crafting our model, \model, for dynamic medical videos, we explore the initial experience baseline for \underline{pipeline design} towards endoscopy video simulation (Fig.~\ref{fig:ppline}). 
We further pioneer the exploration of \underline{experimental baseline} in endoscopic video generation, characterized by the comprehensive collection of clinical videos, and adapt existing generic video generation models for this purpose (Sec.~\ref{exp_set} and Sec.~\ref{exp_compare}). 
Simultaneously, we thoroughly investigate the extensive dimensions of \underline{evaluation baseline} in video generation, including the fidelity of generated content, the improvement in performance for downstream video analysis through data augmentation, and the geometrical quality assessed by multi-view consistency in generation (Sec.~\ref{exp_compare} and Sec.~\ref{exp_study}).

To address the unique challenges of capturing the \textbf{spatio-temporal complexity and fluidity} of real-life medical procedures, we integrate an advanced video transformer architecture with a latent diffusion model, facilitating the extraction of long-range correlations in terms of both spatial and temporal dimension from video data. 
Specially, the training process involves using a pre-trained variational autoencoder~\cite{wang2016auto,li2021unsupervised} to encode video inputs into a latent space. These encoded features are then processed through a sequence of transformer blocks. Furthermore, to ensure consistency across video frames, we employ a feature prior from a 2D foundation model, \dinoC, which helps in regulating key features from different perspectives. Our extensive testing demonstrates that \model~can produce highly realistic endoscopic videos, showcasing its effectiveness and potential for medical video generation with rich dynamics.
In a nutshell, \model~leads the effort in creating complex, high-dimensional surgical video content, setting a benchmark for future medical generative AI research. Our key contributions include:
(\textbf{i})
Introducing a high-fidelity medical video generation framework, tested on endoscopy scenes, laying the groundwork for further advancements in the field.
(\textbf{ii})
Creating the first public benchmark for endoscopy video generation, featuring a comprehensive collection of clinical videos and adapting existing general-purpose generative video models for this purpose.
(\textbf{iii}) Developing a novel technique to infuse generative models with features distilled from a 2D visual foundation model, ensuring consistency and quality across different scales.
(\textbf{iv}) Demonstrating \model's versatility through successful applications in video-based disease diagnosis and 3D surgical scene reconstruction, highlighting its potential for downstream medical tasks.

\section{Method}
We train \model~to generate plausible endoscopy videos given the collection of actual clinical observations.
Firstly, the diffusion backbone is introduced and lifted to handling video formats (Sec.~\ref{sec:ddpm}).
We further introduce cascaded transformer blocks interlaced for spatial and temporal modeling (Sec.~\ref{sec:encoding}).
The prior from image foundation models is distilled to effectively guide the generation of the high-dimensional intricate video representation.

\subsection{
Diffusion Model for Video Generation
}
\label{sec:ddpm}
Generative diffusion models based on Denoising Diffusion Probability Models (DDPM) specializes in transforming disordered noise into desirable samples. By progressively removing noise from Gaussian noise $p\left(\mathbf{x}_T\right)=\mathcal{N}(\mathbf{0}, \mathbf{I})$, these models generate samples aligned with the target data distribution. The forward diffusion step, denoted as $q(\Mat{x}_{t} | \Mat{x}_{t-1})$, adds Gaussian noise into the image $\Mat{x}_t$. The corresponding marginal distribution can be expressed as: $q\left(\Mat{x}_t \mid \Mat{x_0}\right)=\mathcal{N}\left(\alpha_t \Mat{x_0}, \sigma_t^2 \Mat{I}\right)$, where $\alpha_t$ and $\sigma_t$ are designed to converge to $\mathcal{N}(\Mat{0}, \Mat{I})$ when $t$ reaches the end of the forward process~\cite{kingma2021variational,song2020score}.
The reverse diffusion process $p(\Mat{x}_{t-1} | \Mat{x}_t)$ is designed as a noise estimator $\Mat{\epsilon}_\theta(\Mat{x}_t, t)$ that estimates noise from noisy images. The training process involves optimizing the weighted evidence lower bound (ELBO)~\cite{ho2020denoising,kingma2021variational},
\begin{equation}
ELBO=\mean\left[w(t)\left\|\Mat{\epsilon}_\theta\left(\alpha_t \Mat{x}_0+\sigma_t \Mat{\epsilon} ; t\right)-\Mat{\epsilon}\right\|_2^2\right],
\label{eq:ddpm}
\end{equation}
where $\Mat{\epsilon}$ is drawn from $\mathcal{N}(\mathbf{0}, \mathbf{I})$, the timestep $t$ follows a uniform sampling, 
and $w(t)$ serves as a weighting function with $w(t) = 1$.

Lifting diffusion models for videos escalates computational overhead and representation complexity.
Latent Diffusion Models~\cite{rombach2022high} performs the diffusion processes in the encoded latent space rather than the pixel space, improving model efficiency~\cite{grewal2021marketing,li2022knowledge}.
Another strategy~\cite{ho2022video} trains video and image generation simultaneously to improve video generation quality.
We adopt similar strategies in our framework but further introduce new innovations detailed below.

\subsection{
Spatial-temporal Transformer
}
\label{sec:encoding}
Drawing insights from ViT~\cite{liu2021swin} on spatial correlation capture~\cite{sun2022few},  a transformer that exclusively extracts spatial information from tokens that share the same temporal index is introduced as a {Spatial} Transformer.
We employ the patch embedding strategy~\cite{dosovitskiy2021an} to indicate the position embedding for this {Spatial} Transformer.

A {Temporal} Transformer is further introduced to capture temporal information across video frames.
We integrate temporal position embeddings accomplished by using an absolute position encoding strategy, which combines sinusoidal functions of varying frequencies. This strategy enables the model to accurately determine the exact position of each frame within the video sequence.

Specially, given a video clip in the latent space, denoted as ${V} \in \mathbb{R}^{F \times H \times W \times C}$,
where $F$, $H$, $W$, and $C$ denote the number of video frames, height, width, and channel of latent feature maps.
We convert ${V}$ into a sequence of tokens, represented as $\hat{{Z}} \in \mathbb{R}^{N_F \times N_H \times N_W \times D}$.  
The total number of tokens within a video clip in the latent space is  $N_F \times N_H \times N_W$ and $D$ represents the dimension of each token, respectively.
A spatial-temporal positional embedding $\operatorname{PE}$ is integrated into $\hat{{Z}}$. 
Consequently, ${Z} = \hat{{Z}} + PE$ serves as the input for the Transformer backbone.
We reshape ${Z}$ into ${Z^\mathcal{S}} \in \mathbb{R}^{N_F \times L \times D}$ to serve as the input for the Spatial Transformer block, which captures spatial information. Here, $L=N_H \times N_W$ represents the token count of each temporal index. Subsequently, ${Z^\mathcal{S}}$, containing spatial information, is reshaped into ${Z^\mathcal{T}} \in \mathbb{R}^{L \times N_F \times D}$ as the input for Temporal Transformer block, which is used to capture temporal information.
By interlacing a series of Spatial and Temporal Transformers, our model enables modeling long-range spatial correlations and temporal dynamics comprehensively.

\subsection{
Prior-guided Feature Facilitation
}
\label{sec:prior}
Compared to 2D contents, recovering video frames from noise is more difficult, as we cannot adequately approximate the continuity of the temporal dimension, confined to sampling at specific quantized timestamps. 
Hence, it is better to optimize diffusion models to enable the inverse back projection of noise sequences into a latent time-continuous video space aligned with human perception~\cite{skorokhodov2022stylegan,liang2021unsupervised,ZHU2023103230}. 
To achieve this aim, previous work incorporates the power of large language model (LLM)
into the design of video generation~\cite{Zhao_2023_CVPR}, while other approaches effectively capture information using a frozen CLIP encoder~\cite{lin2022frozen}. 
However, they primarily consider semantic correlations between frames and do not adequately address dense correlations (e.g., patches, key points) across frames, which are crucial for frame continuity at a finer granularity. 

Inspired by recent efforts of facilitating dense task~\cite{li2023mask,amir2021deep,jiang2023efficient} via vision foundation models like \dinoC, we consider leveraging the DINO features as they exhibit not only strong semantic correlation but also a potent ability to extract dense correspondence~\cite{amir2021deep}.
We propose to integrate the multi-scale representation~\cite{sinha2020multi,li2022hierarchical} produced by \dino~encoder, ranging from the outputs of its shallow layers to the deeper layers, as a prior obtained by large-scale 2D pre-training to guide the video diffusion training. 
We use attention maps from DINO~\cite{QIU2023109383}, and apply a convolution operation with a stride of 2 and a 3x3 kernel to accommodate them with the dimension of diffusion attention maps. 
Considering the disparate regimes between DINO and \model, we employ the relative distribution similarity~\cite{ekman1969distribution,tang2023all} to match features, by \textit{Pearson correlation},
\begin{equation}
    {Corr}({{A_{\texttt{\model}}}}, {
    \operatorname{Conv}
    (A_{\texttt{DINO}}})
    )
    = \frac{{Cov}
    \left(
    {{A_{\texttt{\model}}}}, {{\operatorname{Conv}
        \left(
        A_{\texttt{DINO}}
        \right)
    }}
    \right)
    } 
    {
    \sqrt{
    {Var}({{A_{\texttt{\model}}}})
    }
    \sqrt{
    {Var}
    \left(
    \operatorname{Conv}
        \left(
        {{A_{\texttt{DINO}}}}
        \right)
    \right)
    \label{eq:prior}
    }
    }
\end{equation}
where $A$ is attention maps, $\operatorname{Conv(\cdot)}$ is the aforementioned convolution layer, 
${Cov}(\cdot)$ is covariance, and ${Var}(\cdot)$ is variance.
We simultaneously maximize this correlation of attention maps at multiple levels (four layers) between DINO encoder and \model~Spatial Transformer blocks, as shown in Fig.~\ref{fig:ppline}. With the guidance of the discriminative DINO prior, the semantic and spatial dependence can be thoroughly enhanced, boosting the photorealism of generated videos. 

Finally, the overall optimization objective for training~\model~is to minimize the following combination of the quantity in Eq.~\ref{eq:ddpm} and Eq.~\ref{eq:prior},
$\mathcal{L}_{DDPM}+ \alpha \mathcal{L}_{Prior} = {ELBO} + \alpha {(1-Corr)}$.
We set $\alpha=0.5$ by grid searching.

\section{Experiments}
\subsection{Experiment Settings}
\label{exp_set}
\noindent \textbf{Datasets and Evaluation.}
We conduct comprehensive experiments on three public endoscopy video datasets: \da~\daCC, \db~\dbCC, \dc~\dcCC.
Following common practice~\cite{ma2024latte}, we extract 16-frame video clips from these datasets using a specific sampling interval, with each frame resized to 128×128 resolution for training.
In the assessment of quantitative comparisons, we employ three evaluation metrics: Fr{\'e}chet Video Distance (FVD) \cite{unterthiner2018towards}, Fr{\'e}chet Inception Distance (FID) \cite{parmar2021buggy}, and Inception Score (IS) \cite{saito2017temporal}. 
Adhering to the evaluation rules in \maC, we compute FVD scores by analyzing 2048 video clips, each comprising 16 frames. 

\noindent\textbf{Implementation Details.}
We employ the AdamW optimizer, with a constant learning rate of $1 \times 10^{-4}$ for training all models. 
We simply apply the basic data augmentation as horizontal flipping. Following standard practices in generative models~\cite{peebles2023scalable,bao2023all}, we use the exponential moving average (EMA) strategy~\cite{tarvainen2017mean,wang2022metateacher,li2021consistent} and report the performance on EMA model for final result sampling.
We directly use the pre-trained variational autoencoder~\cite{ding2022unsupervised,xu2022afsc} from Stable Diffusion~\cite{blattmann2023stable}.
Our model is constructed by $n=28$ Transformer blocks, with a hidden dimension of $d=1152$ with $n=16$ multi-head attention in each block, following ViT~\cite{dosovitskiy2021an}. 

\begin{table}[!t]
  \centering
  \caption{
  {Quantitative Comparisons} on Endoscopic Video Datasets.
  }
    \vspace{0.em}
    \resizebox{\linewidth}{!}
    {
    \begin{tabular}{
    p{4.2cm}>{\centering\arraybackslash}p{0.9cm}>{\centering\arraybackslash}p{0.9cm}>{\centering\arraybackslash}p{0.9cm}>{\centering\arraybackslash}p{0.9cm}>{\centering\arraybackslash}p{0.9cm}>{\centering\arraybackslash}p{0.9cm}>{\centering\arraybackslash}p{0.9cm}>{\centering\arraybackslash}p{0.9cm}>{\centering\arraybackslash}p{0.9cm}
    }
    \toprule
    \multirow{2}[2]{*}{Method} & \multicolumn{3}{c}{\daC} & \multicolumn{3}{c}{\dbC} & \multicolumn{3}{c}{\dcC} \\
    \cmidrule(lr){2-4}   \cmidrule(lr){5-7} \cmidrule(lr){8-10}
                 & FVD\,$\downarrow$ & FID\,$\downarrow$ & IS\,$\uparrow$ & FVD\,$\downarrow$ & FID\,$\downarrow$ & IS\,$\uparrow$ &  FVD\,$\downarrow$ & FID\,$\downarrow$ & IS\,$\uparrow$ \\
    \midrule
    \maC \textcolor{CadetBlue}{(CVPR'22)}&  2110.7   &      226.14       &  2.12  & \cellcolor{top3}183.5          &  \cellcolor{top3}31.61    & \cellcolor{top1}2.77 & \cellcolor{top3}594.1            &  \cellcolor{top3}87.46 & \cellcolor{top3}3.36 \\
    \mcC \textcolor{CadetBlue}{(Arxiv'23)}&        \cellcolor{top3}1036.7      &      \cellcolor{top3}96.85       &    \cellcolor{top3} 1.93          &     1027.8         &        200.9      &  1.46 
 & 1361.5 & 91.25 & 2.65 \\
 \mbC \textcolor{CadetBlue}{(CVPR'23)}&   \cellcolor{top2}468.5       &       \cellcolor{top2}53.17      &   \cellcolor{top2}3.37   & \cellcolor{top2}82.77        &   \cellcolor{top2}17.34          &  \cellcolor{top3}2.53  & \cellcolor{top2}416.2         & \cellcolor{top2}72.87 & \cellcolor{top2}3.56 \\
    \midrule
    \model~(Ours)         &    \cellcolor{top1}{460.7}         &    \cellcolor{top1}13.41         &    \cellcolor{top1}3.90          &    \cellcolor{top1}72.25         &      \cellcolor{top1}10.61      & \cellcolor{top2}2.54 &   \cellcolor{top1}236.2  & \cellcolor{top1}11.20 &  \cellcolor{top1}4.09 \\
    \bottomrule
    \end{tabular}%
    }
  \label{tab_main}%
      \vspace{0.3em}
\end{table}
    
\begin{table}[t!]
\centering
  \caption{{Semi-supervised Classification Result} (F1 Score) on PolyDiag~\cite{tian2022contrastive}. }
      \vspace{0.2em}
      \resizebox{0.72\linewidth}{!}{
    \begin{tabular}{p{2.8cm}>{\centering\arraybackslash}p{3cm}>{\centering\arraybackslash}p{3cm}}
    \toprule
    Method         & Colonoscopic \cite{mesejo2016computer} & CholeTriplet~\cite{nwoye2022rendezvous} \\
    \midrule
       {Supervised-only}    &   74.5& 74.5  \\
    \mcC        &   76.2 (\textcolor{ForestGreen}{\textbf{+}}1.7)       &  78.0 (\textcolor{ForestGreen}{\textbf{+}}3.5) \\
    \midrule
    \model~(Ours)     &   \textbf{87.0} (\textcolor{ForestGreen}{\textbf{+}}\textbf{12.5})     & \textbf{82.0} (\textcolor{ForestGreen}{\textbf{+}}\textbf{7.5}) \\
    \bottomrule
    \end{tabular}
    }
    \label{tab_semi}
        \vspace{-0.2em}
\end{table}

\subsection{Comparison with State-of-the-arts}
\label{exp_compare}
\begin{figure}[t]
 \centering
  \includegraphics[width=0.95\linewidth]{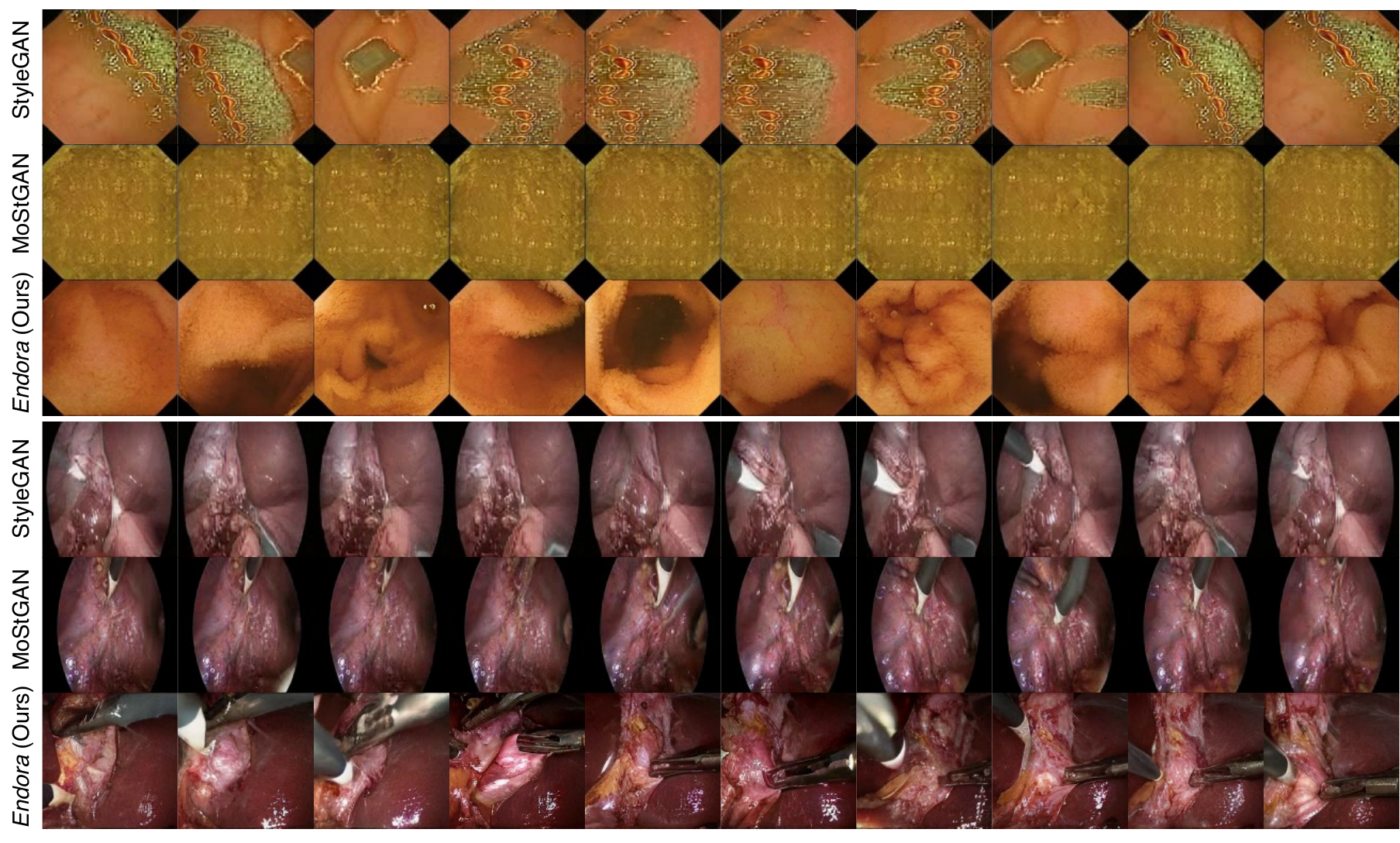}
   \vspace{-0.15em}
 \caption{
  Qualitative Comparison on \db~\dbCC~and~\dcS~\dcCC Datasets.
 }
  \vspace{-0.3em}
 \label{fig_main}
\end{figure}

We conduct performance comparison by replicating several advanced video generation models designed for general scenarios on the endoscopic video datasets, including \maC,~\mbC~and~\mcC.
As shown in Tab.~\ref{tab_main}, \model~excels over the state-of-the-art methods based on GAN~\cite{skorokhodov2022stylegan,shen2023mostgan} for endoscopic video generation in terms of high visual fidelity by all three metrics.
Furthermore, \model~surpasses the advanced diffusion-based method, LVDM~\cite{he2023latent} in all aspects, indicating that \model~effectively generates a accurate video representation of endoscopic scenes.
Fig. \ref{fig_main} further showcases the qualitative results of \model~and prior state-of-the-arts.
We can observe that other techniques result in visually discordant distortions (row 1), restricted content variations (rows 2 and 4), and discontinuous inter-frame transitions (row 5, abrupt intrusion of surgical instruments). In contrast, the video frames generated by \model~(rows 3 and 6) avoid discordant visual distortions (row 1), retain more visual details, and offer superior visual representation of tissues.

\begin{table}[t!]
\centering
\vspace{0.1cm}
  \caption{Ablation Studies of Proposed Components on Colonoscopic~{\daCC} Dataset.}
  \vspace{0.5em}
    \resizebox{0.8\linewidth}{!}{
    \begin{tabular}{>{\centering\arraybackslash}m{1.8cm}>{\centering\arraybackslash}m{2.2cm}>{\centering\arraybackslash}m{1.9cm}>{\centering\arraybackslash}m{1.2cm}>{\centering\arraybackslash}m{1.2cm}>{\centering\arraybackslash}m{1.2cm}
    }
    \toprule
    {Modified Diffusion} & { Spatiotemporal  Encoding} & {Prior Guidance} & {FVD}$\downarrow$ & {FID}$\downarrow$ &{IS}$\uparrow$  \\
    \midrule
            \redcross     &      \redcross        &    \redcross          
            & 611.9 & 22.44 & 3.61 \\
               \gcheck  &     \redcross          &     \redcross         & \multicolumn{1}{c}{593.7} & \multicolumn{1}{c}{17.75} & 3.65\\
               \gcheck  &      \gcheck        &    \redcross          & \multicolumn{1}{c}{493.5} & \multicolumn{1}{c}{13.88} & 3.89 \\
    \midrule
             \gcheck     &     \gcheck         &   \gcheck    & \textbf{460.7}    & \textbf{13.41} & \textbf{3.90}\\
    \bottomrule
    \end{tabular}%
    }
\label{tab_ablation}
    \vspace{-0.3em}
\end{table}

\subsection{Further Empirical Studies}
\label{exp_study}
In this section, we illustrate several potential applications of leveraging the generated videos of our \model~and conduct rigorous ablations on our key strategies.

\noindent\textbf{Case I: \model~as a Temporal Data Augmenter.}
We explore the case of using generated videos as the unlabeled instances for semi-supervised training (by FixMatch~\cite{sohn2020fixmatch}) on the video-based disease diagnosis benchmark (PolyDiag~\cite{tian2022contrastive}).
Specially, we use the randomly selected $n_l=40$ videos in training set of PolyDiag as labeled data, and $n_u=200$ generated videos as the unlabeled data in \daC and \dcC, respectively.
Tab.~\ref{tab_semi} depict the F1 score of disease diagnosis, with the \textcolor{ForestGreen}{\textbf{gain}}
over the baseline using merely labeled training instances labeled ({Supervised-Only}).
The outcomes indicate a notable enhancement in downstream performance attributable to the data generated by \model~compared to not only the {Supervised-only} baseline but also other video generation methodologies, confirming the efficacy of \model~as a reliable video data augmenter for downstream video analysis.

\begin{figure}[!t]
 \centering
  \includegraphics[width=0.95\linewidth]{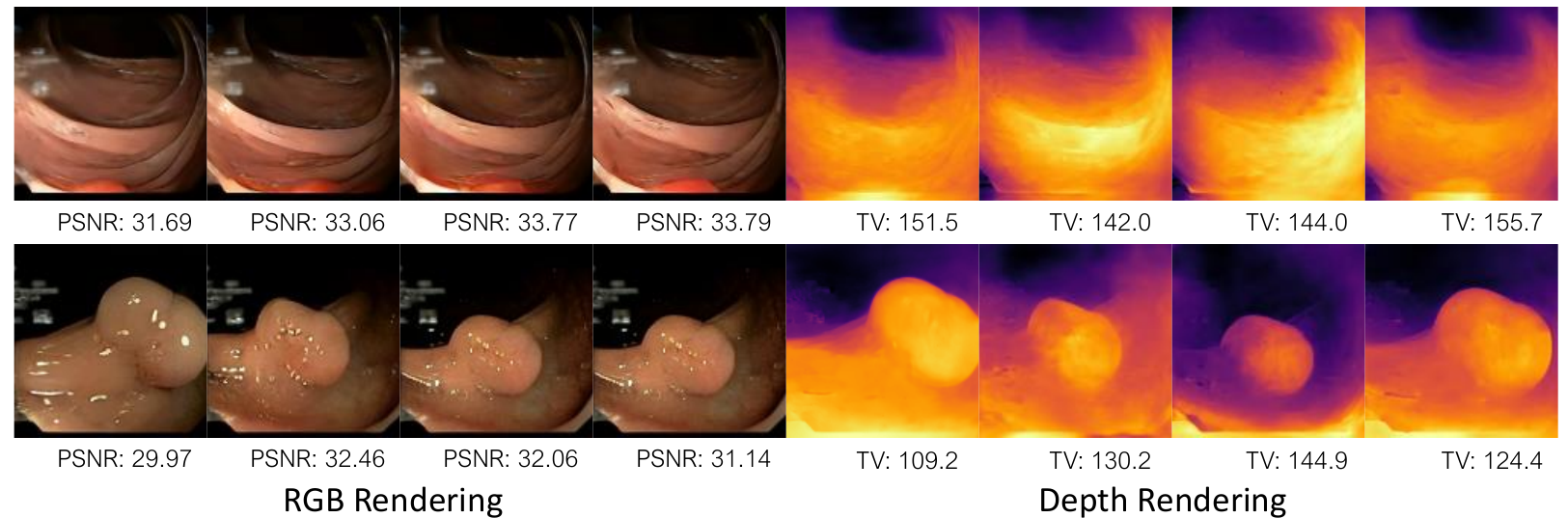}
 \caption{
 RGB and 3D Depth Reconstructed from Generated Videos.
 }
 \vspace{-1em}
 \label{fig_3d}
\end{figure}

\noindent\textbf{Case II: \model~as a Surgical World Simulator.}
Emerging multi-view consistent properties in generated contents~\cite{li2024sora,liu2024sora,li2023steganerf}
inspire our exploration into whether a similar geometric consistency exists in our generated surgical videos.
Specially, from a generated video, we take some frames to as training data for 3D reconstruction (training views), and keep the other frames as testing data (novel views).
We then preprocess the training views with COLMAP~\cite{schoenberger2016sfm,pan2023learning} and then run an off-the-shelf 3D reconstruction pipeline (EndoGaussian~\cite{liu2024endogaussian}) to obtain a reconstructed 3D endoscopy scene. 
Fig.~\ref{fig_3d} provides the visualization of the rendered RGB images and depth maps at novel views, with the image PSNR and depth-wise total variation (TV) labeled.
We can observe that the 3D scenes reconstructed from our generated videos exhibit realistic and continuous geometric structure, showing the potential of \model~to effectively perform surgical world simulation in a multi-view-consistent manner.

\noindent\textbf{Ablation Studies.}
Table~\ref{tab_ablation} presents an ablation study of the key components of the proposed \model. 
Initially, we employ a plain video diffusion model without any of the proposed strategies as a baseline. 
Subsequently, we add the proposed three design strategies one at a time: modified diffusion (Sec.~\ref{sec:ddpm}), spatial-temporal encoding (Sec.~\ref{sec:encoding}), and prior guidance (Sec.~\ref{sec:prior}). 
We can observe they lead to a steady progression in the performance of model, confirming the crucial role of our designed strategies in enhancing the overall efficiency and effectiveness of the endoscopy video generation model.

\section{Conclusion}
This paper introduces \model, a pioneering framework for medical video generation, enabling the creation of high-quality, dynamic, and realistic endoscopy simulation. 
Specifically, we leverage a video transformer to model long-range spatial-temporal relations in videos and employ priors from advanced 2D vision foundation models to enhance the feature extraction.
In rigorous benchmarking experiments of endoscopy videos, \model~delivers superior results in terms of visual quality and impressive potential for downstream video analysis models as data augmenters. 
\model~even enables 3D surgical scene simulation with an off-the-shelf endoscopy reconstruction method.
With \model, we take significant strides towards the application of medical generative AI, providing key insights and setting a strong foundation for future research on medical generated content.

\section{Acknowledgement}
\noindent\textit{Author contributions.}
Conceptualization: C. Li.
Methodology: C. Li, Y. Liu, B. Feng.
Implementation: C. Li, H. Liu, Y. Liu.
Writing: C. Li, B. Feng.
Experiment Design: C. Li, B. Feng, W. Li.
Visualization: C. Li, H. Liu, B. Feng, W. Li.
Supervision: X. Liu, Z. Chen, J. Shao, Y. Yuan.

\appendix
{
   \newpage
        \centering
        \Large
        \textbf{\model: Video Generation Models as \\ Endoscopy Simulators}\\
        \vspace{0.5em}Supplementary Material \\
        \vspace{1.0em}
}

\section{Dataset Setting}

\noindent\textbf{Colonoscopic~\daCC.} The Colonoscopic dataset is a collection of images and video frames sourced from regular colonoscopy procedures, which consists of 210 videos with labels. This dataset is designed to assist in the development of computer-aided diagnosis (CAD) systems for the detection and classification of gastrointestinal lesions observed during colonoscopies.

\noindent\textbf{Kvasir-Capsul~\dbCC.} The Kvasir-Capsule dataset is a publicly available collection of images from capsule endoscopy, which consists of 117 videos with frame-level annotations. The dataset includes various annotated images of different GI tract conditions, normal findings, and common quality issues. It's designed to aid in the development and evaluation of computer-aided diagnostic systems in gastroenterology, providing a valuable resource for training and testing AI models for automatic detection and classification of GI diseases.

\noindent\textbf{CholecTriplet~\dcCC.} CholecTriplet is a specialized collection of endoscopic video data pertaining to laparoscopic cholecystectomy procedures. It comprises 50 videos, with 45 sourced from the Cholec80 dataset and an additional 5 from an internal dataset of the same surgical procedure. These videos are annotated with triplet information in the form (instrument, verb, target), which is essential for identifying the specific actions taken during the surgery.

\section{Hyperparameters}

For all three datasets, we embark on a rigorous training regimen for Endora spanning an extensive $150,000$ epochs, meticulously fine-tuning its parameters with a steadfast learning rate of $1e-4$. This formidable computational endeavor demands the computational prowess of an RTX 6000 boasting a staggering 48GB of memory, tirelessly churning for a span of 3 days. 
The early-stop strategy is adopted to avoid over-fitting~\cite{prechelt2002early,wang2024landa}, which follows the standard practice in video-related tasks.
In our relentless pursuit of enhancing Endora's generative prowess, we judiciously tailor the training parameters. We opt for a localized batch size of $5$, strategically chosen to strike an optimal balance between computational efficiency and model stability. Additionally, to amplify the fidelity of generated outputs, we meticulously set the dimensions of the generated video to $[16, 128, 128]$, wherein $16$ denotes the temporal extent comprising the number of frames, while $[128, 128]$ signifies the spatial resolution of each frame image. Our implementation of Endora is meticulously crafted using the PyTorch framework, harnessing its robust capabilities to drive the model's training and inference pipelines toward paralleled excellence. When imbuing our spatial blocks with prior guidance, we strategically integrate layers 2, 5, 8, and 11 extracted from~\dino~\cite{yao2024cracknex,caron2021emerging}. These layers are meticulously amalgamated with the corresponding spatial components of our blocks, fostering a symbiotic relationship that enriches the guidance received by our model.

\vspace{3em}

\bibliographystyle{splncs04}
\bibliography{ref_full}

\end{document}